%% file: main.tex
\documentclass{article}
\usepackage{ijcai24}

\usepackage{times}
\usepackage{soul}
\usepackage{url}
\usepackage[hidelinks]{hyperref}
\usepackage[utf8]{inputenc}
\usepackage[small]{caption}
\usepackage{graphicx}
\usepackage{amsmath}
\usepackage{amsthm}
\usepackage{booktabs}
\usepackage{algorithm}
\usepackage{algorithmic}
\usepackage[switch]{lineno}

\usepackage{xspace}
\usepackage{multirow}
\usepackage{amssymb}
\usepackage{enumitem}
\usepackage{svg}
\usepackage{subfig}
\usepackage{appendix}

\DeclareMathOperator*{\argmin}{\arg\min}

\newcommand{\our}{\textsc{MUSLA}\xspace}

\newcommand{\dsm}{\textsc{Medium}\xspace}
\newcommand{\dsh}{\textsc{Hard}\xspace}
\newcommand{\minisection}[1]{\vspace{1pt}\noindent\textbf{#1}}

\newcommand{\citet}[1]{\citeauthor{#1}~\shortcite{#1}}

\newcommand\blfootnote[1]{%
  \begingroup
  \renewcommand\thefootnote{}\footnote{#1}%
  \addtocounter{footnote}{-1}%
  \endgroup
}
\title{Looking Ahead to Avoid Being Late: Solving Hard-Constrained Traveling Salesman Problem}

\author{
Jingxiao Chen$^1$
\and
Ziqin Gong$^1$\and
Minghuan Liu$^{1}$\and
Jun Wang$^2$\and
Yong Yu$^1$\And
Weinan Zhang$^1$
\affiliations
$^1$Shanghai Jiao Tong University\\
$^2$University College London\\
\emails
\{timemachine, gongzq0301, minghuanliu\}@sjtu.edu.cn,
jun.wang@cs.ucl.ac.uk,\\
\{yyu, wnzhang\}@sjtu.edu.cn
}
\begin{document}

\maketitle

\begin{abstract}
  Many real-world problems can be formulated as a constrained Traveling Salesman Problem (TSP). However, the constraints are always complex and numerous, making the TSPs challenging to solve.
  When the number of complicated constraints grows, it is time-consuming for traditional heuristic algorithms to avoid illegitimate outcomes.  Learning-based methods provide an alternative to solve TSPs in a soft manner, which also supports GPU acceleration to generate solutions quickly.
  Nevertheless, the soft manner inevitably results in difficulty solving hard-constrained problems with learning algorithms, and the conflicts between legality and optimality may substantially affect the optimality of the solution. 
  To overcome this problem and to have an effective solution against hard constraints, we proposed a novel learning-based method that uses looking-ahead information as the feature to improve the legality of TSP with Time Windows (TSPTW) solutions.
  Besides, we constructed TSPTW datasets with hard constraints in order to accurately evaluate and benchmark the statistical performance of various approaches, which can serve the community for future research.
  With comprehensive experiments on diverse datasets, 
  \our outperforms existing baselines and shows generalizability potential.
  \blfootnote{Preliminary work. Under review.}
\end{abstract}

% 1 page
\input{Introduction}
% 1 page
\input{RelatedWork}
% 1 page
\input{Problem}
% 2 pages
\input{Methods}
% 1 page
\input{Dataset}
% 2.5 pages
\input{Experiments}
% 1 page
\input{Conclusion}

% \resetlinenumber

% \bibliographystyle{named}
% \bibliography{aaai24}

\input{ref}
% \newpage
\clearpage

\appendix
\setcounter{page}{1}
\input{appendix}
\input{supplement}

\end{document}

%% file: Introduction.tex
\section{Introduction}
% 1. intro the problem: real-world, constraints, TSP-TW
NP-hard combinatorial optimization problems play a vital role in modern practical applications and industries. In the real world, problems are always attached to a set of complex constraints, making them challenging to solve. 
Typically, the constraints in combinatorial optimization problems include hard and soft ones. 
% Hard constraints strictly require conditions on feasible solutions, while soft constraints tolerate subtle violations. 
Soft constraints tolerate slightly violating the constraints in a small range, whereas hard constraints strictly prohibit any violation. 
% For example, in the soft-constrained problem, Capacitated Vehicle Routing Problem (CVRP), simply allocating enough vehicles definitely guarantee the capacity of goods demanded by each city. Whereas in the hard-constrained Travelling Salesman Problems with time windows (TSPTW), the shortest path is obviously not guaranteed to satisfy the time-window constraint, and a longer path may also cause more violations.

In this paper, we focused on the set of popular Traveling Salesman Problems (TSPs), which are described as asking a salesman to visit each city with minimizing the total length of the tour. This scenario is commonly used in profit optimization in industrial production procedures. 
Considering different kinds of constraints, there are many variants of TSPs.
For instance, Traveling Salesman Problems with time windows (TSPTW) is a famous hard-constrained variant of TSP in the vehicle routing problem (VRP) family, which constrains the salesman from visiting each city in a particular range of time and meanwhile minimizes the total length of the tour; 
In comparison, the capacitated vehicle routing problem (CVRP) puts soft constraints on TSP which require multiple salesmen, also called vehicles, with limited carrying capacity to deliver items to various locations. 
% Other hard-constrained variants of TSP are the Pickup-and-delivery traveling salesman problem (PDTSP).

% 2. Machine Learning, RL approaches, GNN, Attention
% With the development of machine learning (ML) and reinforcement learning (RL), an increasing number of recent works focus on solving combinatorial optimization problems including TSP with ML or RL methods. For problems with the graph structure, Graph neural network was widely used with ....
% Attention structure ... 

% 3. Motivations: Different between Heuristics and learning-based model
% Ours Adv:
% Learning vs Search:
%   - faster: get result in a few trails, end-to-end, can be accelerated by GPU
% SL vs RL:
%   - need reward shaping: drop in a local optimal
%   - time-consuming: need to sample tons of data with env
% SL-ours vs RL:
%   - only precompile limited expert datas
% SL-ours vs SL-old:
%   - lacking illegal signal

% Learning vs Search
Traditional heuristic searching approaches, such as LKH3~\cite{Helsgaun2017AnEO}, traverse a large number of solutions and search for the best one that satisfies the constraints. The Searching method guarantees finding feasible solutions under arbitrary constraints. 
However, searching-based algorithms require a well-designed heuristic function for a specific problem and consume a lot of time to search for different solutions for a new problem instance. 
In order to improve efficiency, recent work has turned to machine learning techniques that construct end-to-end solvers to generate high-quality solutions in a few trials and allow GPU acceleration to improve efficiency further.
However, limited trials make it difficult for these solvers to find a feasible solution. Therefore, solving hard constraints with learning-based approaches becomes more challenging. 
% Are the challenges of the Learning-based method described clearly?

% 4. Emphasize hard constrains chellange in ML
% SL vs RL:
In order to deal with hard constraints, most learning-based methods train an end-to-end solver in the RL paradigm and 
% TSP and its variants are modeled as decision-making problems with RL framework\cite{Kool2018AttentionLT, Falkner2020LearningTS}.
relax the time windows as soft constraints~\cite{tang2022learning} or solve a soft-constrained variant of TSPTW, \textit{e.g.}, Traveling Salesman Problem with Time Windows and Rejections~\cite{Zhang2020DeepRL}. 
In contrast to RL methods requiring elaborate reward designs and millions of environment interactions, 
supervised learning (SL) can be easily trained with fixed offline expert datasets, which is more practical for real-world problems. However, SL was only used to solve regular TSPs in earlier research~\cite{Vinyals2015PointerN,Joshi2019AnEG,Nowak2017ANO,Xing2020AGN}, but it is rarely used for constrained ones.
Challenges, such as obtaining information about constraint boundaries from expert data, prevent applying the SL method to hard-constrained TSPs.
% 5. Additional chellange in SL paradiam: no valid signal, and intro Ours
% advantage vs RL: 1. no need for designing reward
%                  2. faster converge, do not need to sample with environments
%                  3. genralize to different tasks
% advantage vs Heuristic: faster

In this paper, we proposed a novel algorithm named \underline{MU}lti-\underline{S}tep \underline{L}ook-\underline{A}head (\our) to efficiently resolve one of the hard-constrained TSPs, TSPTW. 
In detail, \our introduced a novel one-step look-ahead way to gather future information regarding constraint boundaries and train a policy $\pi^{+1}$ by imitating an expert. Based on the well-trained policy, we augment expert datasets with multi-step look-ahead information collection.
The gathered information about future situations provides a better perception of the constraint boundaries. With the enhanced dataset, we further train the \our policy $\pi^{+m}$.

In a nutshell, our main technical contributions are threefold:
\begin{itemize}[leftmargin=15pt]
    \item We propose \our, which includes a novel looking-ahead mechanism based on supervised learning methods. \our enhanced optimality and legality of SL solutions by augmenting datasets with searched information.
    % We demonstrate that the legality of supervised learning approaches can be enhanced by the looking-ahead mechanism. 
    % by incorporating information about constraint boundaries.
    \item We design two kinds of TSPTW datasets to evaluate the performance of solvers better.
    \item Compared to state-of-the-art work with RL paradigm, \our outperforms other baselines and has a good balance between solution quality and validation rates.

\end{itemize}

%% file: RelatedWork.tex
\section{Related Work}
% Neural Networks
% Reinforcement Learning
% Supervised Learning methods
% 
% \jingxiao{TODO}
% learning-based TSP (RL)
% TSPTW-RL + other TSP
% SL + Heuristic Algorithms
% Discuss the adv vs RL

% SL vs RL:
Recent learning-based work solved TSP and its variants in a reinforcement learning paradigm. 
Graph Neural Networks~\cite{Kipf2016SemiSupervisedCW,Velickovic2017GraphAN,Battaglia2018RelationalIB} and Attention mechanism~\cite{Vaswani2017AttentionIA} are the major architectures in state-of-the-art work. 
\citet{Kwon2020POMOPO,Kim2022SymNCOLS}  leverage symmetricities to improve the generalization capability of learning-based solver.
Recent work~\cite{Velickovic2017GraphAN} has achieved comparable performance to traditional methods on a scale of no more than 200 nodes. On large-scaled TSPs, learning-based methods even demonstrated faster execution with slightly dropped performance~\cite{Fu2020GeneralizeAS}. In the RL architecture, the learning-based policy trials and errors in sampled problem instances and updates the policy according to reward signals. The reward function is usually set to be the negation of the tour length.

% TSP and its variants are modeled as decision-making problems with RL framework\cite{Kool2018AttentionLT, Falkner2020LearningTS}.
% Due to the limitation of the learning-based algorithm, learning-based works 
For constrained TSP, such as Traveling Salesman
Problem with Time Windows, solving constraints is treated as another optimization objective, which brings challenges to learning-based methods.
\citet{Ma2019CombinatorialOB} proposed a hierarchical RL policy to separate constraints from original optimization objectives. \citet{Alharbi2021SolvingTS} used a complex hybrid network architecture to improve solution quality.
Another feasible option is to solve reductions of TSPTW.
\citet{tang2022learning} relaxed time windows as soft constraints, and \citet{Zhang2020DeepRL} solved a soft-constrained variant of TSPTW. Other methods overcome the challenge of constraints by combining learning-based methods with traditional algorithms~\cite{Cappart2020CombiningRL,Zheng2022ReinforcedLA,Papalitsas2015InitializationMF}, but at the cost of the computational burden. 
% In order to solve constraints, the violation degree is set as an additional objective, and the algorithm improves itself with millions of trials and errors in the training process. 
For RL approaches, there are usually two drawbacks to solving constrained TSP. One is the additional objective requires well-designed reward shaping to balance legality and optimality. The other is that RL methods require a lot of trial-and-error and interactions with the environment during training, resulting in high training costs especially for real-world problems.

% 5. Additional chellange in SL paradiam: no valid signal, and intro Ours
% advantage vs RL: 1. no need for designing reward
%                  2. faster converge, do not need to sample with environments
%                  3. genralize to different tasks
% advantage vs Heuristic: faster

Supervised learning (SL) is an alternative  learning-based method to solve TSP~\cite{Vinyals2015PointerN,Joshi2019AnEG,Nowak2017ANO,Xing2020AGN}. The SL policy is trained on datasets labeled by experts, thus avoiding additional trial and error and reward design. Recent work usually uses exact algorithm~\cite{Concorde2006,Ortools2023} and heuristic algorithms~\cite{Helsgaun2017AnEO} as oracles to generate the datasets. 
% However, it is rare to use the SL algorithm to solve constrained TSP. 

%% file: Problem.tex
\section{Problem Formulation}
% In this section, we introduce the traveling salesman problem with time windows (TSPTW) and describe the challenge in a hard-constrained setting.
\subsection{Traveling Salesman Problem with Time Windows}
We first introduce the traveling salesman problem with time windows (TSPTW) and describe the challenge in a hard-constrained setting.
Consider a symmetric complete graph $\mathcal{G} = (V, E)$, where the set of nodes $V=\{0, 1, \dots, n\} $ denotes the set of \textit{cities}, and the set of edges $ E \subset V \times V$ denotes the set of \textit{paths} between every two \textit{cities}. Each \textit{city} $i \in V$ has two attributes, the 2-D coordination $a_i$ and the visiting time window $[t^s_i, t^e_i]$. The length of edge $e_{i,j} \in E$ is $L_{i,j}=||a_i-a_j||_2$, where $|| \cdot ||_2$ is the $l_2$ norm. 
The goal of TSPTW is to minimize the distance of the total path by asking the \textit{salesman} to visit all the \textit{cities} exactly once and return to the start \textit{city} under the constraint that each \textit{city} must be visited within the given time window. A formal definition is as follows:
\begin{equation}
\begin{aligned}
\min_{X=\{x_0, x_1, x_2, \dots, x_n\} } & L_{x_n, x_{0}} + \sum_{i=0}^{n-1} L_{x_i, x_{i+1}} \\
\textit{s.t.}\  & t^s_{x_i} \leq t_i \leq t^e_{x_i} ,\\
\end{aligned}
\end{equation}
where $X$ is a solution tour, and $t_i$ is the time to visit city $x_i$. 
For simplicity, we presume the speed number is equal to $1$ and exclude the speed element from the calculation.
A legal $X$ is a permutation of nodes, indicating $x_i \neq x_{i'}, \forall i\neq i'$.
Without loss of generality, we assume $x_0=0$ and traveling time between $x_i, x_{i+1}$ is $L_{x_i, x_{i+1}}$. 

A hard constraint is one that must be satisfied at all times. It is notable that $t_i \leq t^e_{x_i}$ is a hard constraint, while $t_i \geq t^s_{x_i}$ is not. If city $x_i$ is visited before the earliest time constraint $t^s_{x_i}$, the salesman should wait until $t^s_{x_i}$, \textit{i.e.}, $t_{i} = \max\{t_{i-1} + L_{x_{i-1},x_{i}} , t^s_{x_i}\}$.

The goal of the TSPTW could also be minimizing the total tour duration, also called make-span, rather than the distance of the tour~\cite{Kara2015FormulationsFM}. 
However, optimizing toward the make-span optimality may also reduce violations of constraints. It is difficult to design a reasonable evaluation dataset with the goal of make-span to clarify the model's ability to balance optimality and legality, which makes us ignore this situation.

% why we use decision making

As a hard-constrained problem, TSPTW is commonly regarded as a multi-objective problem in current learning-based paradigms. The optimal objective minimizes the tour distance and the legal objective minimizes the violation of constraints. In contrast to the soft-constrained problem where legality is considered a secondary optimization objective, optimality and legality should share the same status in a hard-constrained setting. Different from other constrained combinatorial optimization problems, such as CVRP and Vehicle Routing Problem with Time
Windows (VRPTW), the legality of the solution in TSPTW can not be satisfied by feasible masking.

\subsection{Supervised Learning with Route Construction}
In recent learning-based work, solutions of TSP and its variants are modeled as an $n$-step route construction process~\cite{Kool2018AttentionLT}. Similar to the process of salesman travel, the solution sequence $X$ is generated step by step in order with a learning-based policy $\pi_\theta$. Given the current partial tour $X_{0:i}=\{x_0, \dots, x_i\}$ and property of problem instance $g$, policy $\pi_\theta$ captures $p(x'|X_{0:i}, g)$, \textit{i.e.}, the probability of visiting the next node $x_i$ at step $i$. After $n$ steps of route construction, the solution tour $X=\{x_0, x_1, \dots, x_n\}$ is given by $p(X|g) = \prod_{i=0}^{n-1} p(x_{i+1} | X_{0:i},g)$. 

In order to train the policy $\pi_\theta$, an expert dataset $\mathcal{D}$ is required, which includes multiple pairs of TSPTW instances $g$ and expert solutions $X^*$. For each problem instance $g\sim \mathcal{D}$, the expert solution $X^*$ is generated by a high-quality expert solver $X^*=\pi_\text{expert}(g)$. Hence the policy is trained to imitate the expert policy, whose objective, written formally, is maximum likelihood estimation:
\begin{equation}
J(\theta) = \max_\theta \log p_{\pi_\theta}(X=X^* | g).
\end{equation}

%% file: Methods.tex
\section{ Methodology } \label{sec:method}

\begin{figure}[t]
  \centering
  \includegraphics[width=\linewidth]{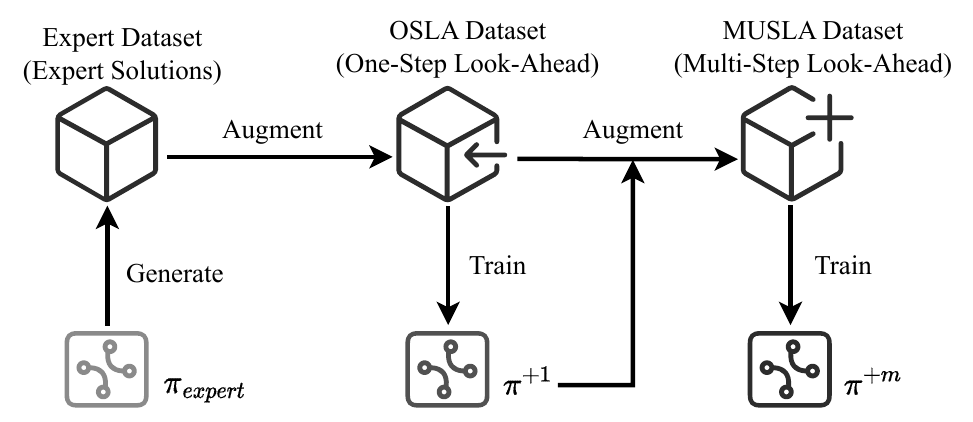}
  \vspace{-1em}
  \caption{Method pipeline of \our. We labeled expert datasets with LKH3 solutions in order to train a faster learning-based solver. With one-step look-ahead augmented datasets, we trained OSLA policy $\pi^{+1}$. OSLA policy directed further multi-step look-ahead data augmentation, resulting in MUSLA policy $\pi^{+m}$.}\label{fig:method_pipeline}
  \vspace{-1em}
\end{figure}

In this section, we introduce our looking-ahead method for TSPTW. The pipeline of training is visualized in Figure~\ref{fig:method_pipeline}. 
%At first, we construct datasets following section~\ref{sec:dataset} and collect expert solutions with LKH3~\cite{Helsgaun2017AnEO}. 
At first, we redesign dynamic features for TSPTW and add a dynamic encoder module for our SL model.
Then, we augment the expert dataset with the one-step look-ahead mechanism and train a supervised learning policy $\pi_\theta^{+1}$.
Utilizing the policy, we gathered multi-step look-ahead (\our) information to refine the expert datasets further and train the \our policy $\pi_\theta^{+m}$. 
Finally, we introduce a technique that can better adapt \our policy to specific problem instances at inference time by modifying dynamic information.

Our method employs supervised learning instead of reinforcement learning due to two primary factors. At first, RL requires additional reward shaping to balance the optimality and legality of solutions. Secondly, applying \our to RL necessitates the repeated collection of augmented information throughout the entire learning process, resulting in an unacceptable computing burden.

\subsection{Learning with Dynamic Information}\label{sec:method1}
We consider a supervised learned policy $\pi_\theta$ using existing expert datasets, which consists of an encoder and a decoder. The encoder receives path-building information and generates embedding, and the decoder takes embedding as input and generates subsequent nodes.
Similar to \citet{Alharbi2021SolvingTS}, $\pi_\theta$ takes static information $I^s(g) = \{a_i, t^s_i, t^e_i\}$ and history embedding $I^h(X_{0:i})$ of the current tour $X_{0:i}$ as input. 
However, this does not describe each possible next node $x'$ on current solution $X_{0:i}$ very well. It may cause cumulative errors in the time and distance for the next step selection, resulting in incorrect estimation for time window constraints.
To alleviate this challenge, we design an additional dynamic node feature $I^d(X_{0:i}, x', g)$. 
In particular, the dynamic feature of each unvisited node $x' \in V \setminus X_{0:i}$ includes differences between $x_i$ and $x'$ in location and time dimensions. For example, in location dimension, features $\{L_{x', x_i}, a_{x'}-a_{x_i}\}$ describe the distance and direction between $x'$ and current node $x_i$. In time dimension, features $\{t^s_{x'}-t_i, t^e_{x'}-t_i\}$ describe the time difference between time windows of $x'$ and $x_i$. 
With dynamic information, visiting probability of the next node $x'$ can be written as $p(x=x' | X_{0:i}, g)=\pi_\theta(x', I^s(g), I^h(X_{0:i}), I^d(X_{0:i}, x', g))$.
In the subsequent representations, we streamline $I^s$ and $I^h$, resulting in the simplification of the policy as $\pi_\theta(x', I^d)$.
Details of feature design and model architecture are further described in supplementary materials.
% Appendix~\ref{sec:app_feature}.

\begin{figure*}
  \centering
  % \includesvg[width=0.8\linewidth]{musla.svg}
  \includegraphics[width=0.65\linewidth]{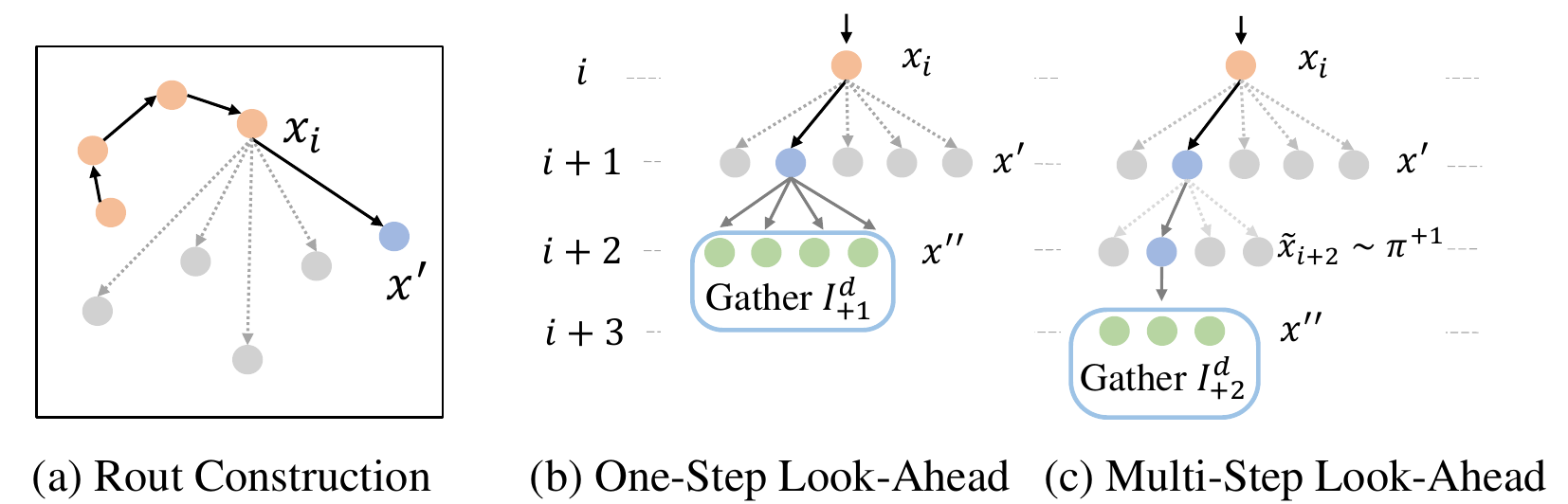}
  % \fbox{\rule[-.5cm]{0cm}{4cm} \rule[-.5cm]{4cm}{0cm}}
  \caption{Illustration of the multi-step look-ahead mechanism for $m=1$. Subfigure (a) shows the route construction at step $i$. Subfigures (b) and (c) illustrate the process of information gathering. Orange nodes have been determined to be in the current route $X_{0:i}$. Blue nodes are temporarily added to the route during the search. Future information is gathered from green points.} \label{fig:look-ahead}
\end{figure*}

\subsection{Dataset Augmentation}
Different from soft-constrained settings, violations of hard constraints result in complete failure. Hence, the major challenge of solving a hard-constrained TSPTW with the supervised learning policy $\pi_\theta$ is to learn non-existing information about constraint boundaries. 
Since only the legal solutions are preserved in expert datasets, the learned policy may lack sufficient information to determine when the time constraints are violated. 
To provide constraint information and train a robust policy, our solution is to augment datasets using a look-ahead mechanism. Depending on the expanded step, we introduce the one-step look-ahead mechanism, based on which we further propose the multi-step look-ahead mechanism.

Figure~\ref{fig:look-ahead} presents a schematic diagram of our approach. To ascertain if node $x'$ should be the next node, we can enumerate all feasible routes that include $x'$ by brute force. However, this approach is obviously impractical given the vast number of potential solutions. Hence, based on node $x'$, we try $m$ subsequent steps to construct an illusory partial solution with a length of $i+1+m$. By exploring several illusory partial solutions of $x'$, we gather $m$-step look-ahead information $I^d_{+m}$ as the policy's input.
% to influence the policy's decision on whether x' should be the subsequent location.
% We follow the route construction process to generate solutions. At $i$-th step, given the current solution $X_{0:i}$, we gather look-ahead features $I^d_{+m}$ for node $x'$ by traverse multiple illusory partial solutions $X'=\{x_0, \dots, x_i, x', \tilde x_{i+2}, \dots, \tilde x_{i+1+m}\}$. 

\subsubsection{One-Step Look-Ahead} \label{sec:method2}

We start from the one-step look-ahead (OSLA) mechanism.
OSLA expands information on each node by trying to construct solutions one step ahead.
At step $i$, we iterate each unvisited node $x'$ to construct an illusory $i+1$ partial solution $X' = \{x_0, \dots, x_i, x'\}$ and gather information for all unvisited node $x'' \in V \setminus X' $ based on the imaginary $i+1$ step. The gathered information $I^{d}_{+1}(X',g)$ is used as additional dynamic features of node $x'$ and helps determine the probability of choosing $x'$ at the current step.

More specifically, the OSLA information includes two types of value. In the first type of feature, we select the set of nodes that are already to be late, i.e. $X''_\text{late}=\{x''\in V\setminus X' | t_{i+1} + L_{x', x''} > t^e_{x''}  \}$. We used the number of late nodes $|X''_\text{late}|$ and the maximum late time as features to capture the constraint violations of selecting $x'$.
In the second type of feature, assuming $X''_\text{late}=\varnothing$, we add  distance and time overhead when greedily visiting node $x''$ with minimum time overhead.
% $\argmin_{x''} \{\max\{t_{i+1} + L_{x', x''}, t^s_{x''}\}\}$

With the OSLA information as additional input, we can train an OSLA policy $\pi^{+1}_\theta(x', I^d, I^d_{+1})$ by supervised learning which is able to learn constraint boundaries by possible timeouts in the future. It is notable that future information is gathered based on expert solutions and does not impose an additional burden on the training process. 
% Once pre-gathering OSLA information, it can be used to train any different supervised learning policy. 

\subsubsection{Multi-Step Look-Ahead with An OSLA Policy} \label{sec:method3}
Searching with one step helps to know about the constraint boundaries, but the collected information is still limited. The effect of a choice may require successive attempts to judge. To gather more guidance information from the future at an acceptable computational cost, we propose a multi-step look-ahead (\our) mechanism to further augment expert datasets.

As the number of steps increases, the number of potential solutions increases geometrically. In order to ensure the feasibility of calculating, we leveraged a pre-trained OSLA policy to screen for meaningful choices. Different from one-step augmentation, we only gather future information for the top $k$ nodes $x'$ with the highest probability of expert selection. Here we use policy $\pi_\theta^{+1}$ as an approximation to the expert strategy. For a specific node $x'$, we continually construct $m$ steps to get an illusory $i+1+m$ partial solution $\Tilde{X}' = \{x_0, \dots, x_i, x', \Tilde{x}_{i+2}, \dots, \Tilde{x}_{i+1+m}\}$. 
After augmentation, we also gather information $I^d_{+2}(\Tilde{X}',g)$ for all unvisited node $\Tilde{x}'' \in V \setminus \Tilde{X}'$ as the \our features of $x'$. 

In this paper, we used a hyper-parameter setting of $k=5, m=1$ to gather multi-step features. 
Augmenting expert datasets with dynamic features $I^d$, $I^d_{+1}$and $I^d_{+2}$, we train the \our policy $\pi_\theta^{+2}(x', I^d, I^d_{+1} I^d_{+2})$ on these data. 

\subsection{Trade-off between Optimality and Legality} \label{sec:method4}
After training on augmented expert datasets, we obtain a robust \our policy that balances optimality and legality. However, by modifying dynamic features, we can still adjust the balance of the two goals and adapt to specific TSPTW instances at the inference stage.
For example, for a problem instance with tight time windows, the model might give an illegal tour that times out slightly. By modifying the tour time $t_i$ with $t'_i=t_i + \epsilon,\ \epsilon>0$, the dynamic feature of the remaining time to reach a node $t^e_{x'} - t'_i$ becomes smaller, so the model adopts a more conservative strategy to avoid timeouts. Following this idea, we extend \our to \our-adapt. \our-adapt tries different time offset values $\epsilon$ during inference and chooses the optimal legal solution as the final solution.

%% file: Dataset.tex
\section{Datasets with Hard-Constraints} \label{sec:dataset}

In order to highlight the ability of the algorithm to balance optimality and legality, we create two kinds of TSPTW datasets, \dsm and \dsh. \dsm is designed with a clear random distribution that increases the difficulty of satisfying the time window constraints. \dsh is constructed in a complex way and aims to evaluate the generalization ability of the model. The expert solutions of our datasets are given by LKH3 solver~\cite{Helsgaun2017AnEO} and we only discuss the generation of problem instances $g\sim D$ in this section.

% Principles for design
Traditional approaches, like LKH3, provide datasets with a limited number of problem instances, which is insufficient for training learning-based algorithms.
Most recent learning-based TSPTW work did not follow a uniform way to generate data and ignored the importance of dataset quality. \citet{Cappart2020CombiningRL} and \citet{Ma2019CombinatorialOB} generated time windows following the visiting time $t'_i$ of a given solution $X'$. Their datasets give too strong prior assumptions about the time window and also oversimplify this problem. A trivial greedy policy that takes unvisited node $i$ with the smallest value $t^s_{x_i}$ is able to obtain a near-optimal solution with low violation of constraints. \citet{Zhang2020DeepRL} generated datasets with pure randomization, but inappropriate parameter settings lead to the absence of legal solutions for most problem instances. 

We demonstrate detailed analysis in Appendix~\ref{sec:app_weak_data}.
According to the backward of previous works, we propose datasets with proper constraints to highlight the challenge of balancing two objectives.

% Dataset: Random
\paragraph{Medium dataset.} 
Similar to \citet{Zhang2020DeepRL}, we generate the random dataset \dsm by randomly sampling coordinates $a_i$ , time windows $[t^s_i, t^e_i]$ of $n+1$ nodes. The 2D-coordinates $a_i$ are sampled uniformly in a grid of $\mathcal U[0, 100]^2$. The time windows are given by sampling start time $t^s_i$ and width $t^e_i-t^s_i$ uniformly:
$$ t^s_i \sim \mathcal{U} [0, T_{n}], \ t^e_i = t^s_i + T_{n} \cdot \mathcal{U}[ \alpha, \beta] $$
where $\alpha, \beta$ are hyper-parameters, and $T_{n}$ is the expected distance of an arbitrary TSP tour on $n+1$ nodes. For $n = 20$, a rough estimate of $T_n$ is $T_{20} \approx 10.9$. By expanding the sampling range of the start time, the size of the feasible solution set in the dataset can be effectively limited, thus bringing conflicts between optimality and legality. 

% Dataset: Construction
\paragraph{Hard dataset.}
In order to guarantee the credibility of evaluation, we add a more difficult dataset \dsh. In this dataset, we sample problem instances for training and evaluation from slightly different distributions. This setting is similar to the real-world situation, and there is often a certain deviation between the training and test scenarios. We test the generalization ability of the model to different problem instances on this dataset.

We provide a detailed description and pseudocode of these constructed datasets in supplementary materials.
% Appendix~\ref{sec:app_data}.

\paragraph{Supplementary training dataset.} \label{sec:sup_dataset}
Due to the incapacity to iterate through different problem instances in TSPTW, a learning-based algorithm trained on fixed datasets tends to overfit or be unstable when solving new problems. To alleviate the issue, we augment variants of the aforementioned datasets to stabilize the model performance. The supplementary dataset includes simplified datasets, such as removing one or two sides of time windows, and also complex datasets. For the purpose of elucidating our conclusion, we trained our model on all dataset variants but only presented evaluation results for Easy and Hard datasets. Details regarding additional datasets are provided in supplementary materials.
% Appendix~\ref{sec:app_data}.
% Current learning-based works for TSPTW are based on training and evaluating datasets with fixed problem sizes. 

% In a hard-constrained problem, the 
% Given an arbitrary TSPTW dataset to evaluate the performance of algorithms, the ability to balance optimality and legality of solutions can not be 

% principle 1: time windows information 
% principle 2: time window should not generate on any particular solutions

%% file: Experiments.tex
\begin{table*}[htb]
\centering
\caption{The result table compares the performance of our model with other baselines. } \label{table:results}
\resizebox{0.8\linewidth}{!}{
\begin{tabular}{c|c|ccc|ccc|c}  
\toprule
 \multicolumn{2}{c|}{\multirow{2}{*}{Methods}} & \multicolumn{3}{c|}{\dsm} & \multicolumn{3}{c|}{\dsh} & \multirow{2}{*}{Time(s)} \\
 \cmidrule{3-8}
 \multicolumn{2}{c|}{}  & Illegal(\%) & Gap(\%)  & \multicolumn{1}{l|}{Timeout} & Illegal(\%) & Gap(\%) & \multicolumn{1}{l|}{Timeout} &  \\ 
 \midrule
\multirow{7}{*}{$N=20$} & LKH3  & 0.00 & 0.00 & 0.00 & 0.00 & 0.00 & 0.00 &  0.2 \\ %0.30/0.10
        ~ & Greedy-MT  & 0.00 & 95.97 & 0.00 & 12.50 & 51.70 & 7.48 & 0.27 \\ 
        ~ & Greedy-LT  & 0.00 & 128.82 & 0.00 & 5.13 & 168.57 & 4.70 & 0.08 \\ 
        ~ & AM  & 5.34 & 16.22  & 0.03  & 83.00 & 52.13  & 5.15  & 0.28 \\ 
        ~ & JAMPR  & 0.00 & 116.02 & 0.00 & 8.29 & 74.31 & 15.56 & 6.03 \\ 
        ~ & \our $\pi^{+2}$ & 3.73  & \textbf{5.33}  & 0.26  & 5.20  & \textbf{12.10}  & 0.26  & 3.30   \\ 
        ~ & \our adapt & 0.20  & \textbf{4.02}  & 0.24  & 0.40  & \textbf{10.25}  & 0.10  & 19.72  \\ \midrule
 \multirow{7}{*}{$N=50$} & LKH3  & 0.00 & 0.00 & 0.00 & 0.00 & 0.00 & 0.00 & 11.64 \\ %17.56, 5.72
        ~ & Greedy-MT  & 0.00 & 196.12 & 0.00 & 18.64 & 69.08 & 25.33 & 0.27 \\ 
        ~ & Greedy-LT  & 0.00 & 257.49 & 0.00 & 17.19 & 311.07 & 89.21 & 0.08 \\ 
        ~ & AM  & 9.90  & 32.68  & 0.09  & 49.50  & 65.88  & 9.08 & 0.27  \\ 
        ~ & JAMPR  & 0.00 & 249.03 & 0.00 & 1.31 & 207.10 & 0.88 & 7.30 \\ 
        ~ & \our $\pi^{+2}$  & 8.20  & \textbf{7.32}  & 1.80  & 18.90  & \textbf{16.71}  & 4.42  & 7.63   \\ 
        ~ & \our adapt & 0.10  & \textbf{5.63}  & 0.99  & 3.10  & \textbf{15.24}  & 2.22  & 45.97  \\ 
        \midrule
  \multirow{7}{*}{$N=100$} & LKH3  & 0.00 & 0.00 & 0.00 & 0.00 & 0.00 & 0.00 & 7588.75 \\ 
        ~ & Greedy-MT  & 0.00 & 314.04 & 0.00 & 20.42 & 79.25 & 36.03 & 0.30 \\ 
        ~ & Greedy-LT  & 0.00 & 409.62 & 0.00 & 30.02 & 468.76 & 408.18 & 0.08 \\ 
        ~ & AM & 9.00  & 239.57  & 0.05  & 33.40  & 132.18  & 3.42  & 3.01 \\ 
        ~ & JAMPR  & 100.00 & N/A & 14.82 & 100.00 & N/A & 734.44 & 9.53 \\
        ~ & \our $\pi^{+2}$  & 18.60  & \textbf{14.60}  & 24.81  & 50.50  & \textbf{37.05}  & 96.39   & 58.83   \\ 
        ~ & \our adapt & 0.60  & \textbf{12.01}  & 9.59  & 31.90  & \textbf{35.59}  & 89.57  & 403.53 \\
        % ~ & dpdp (bs 5k) & 0 & 4.59 & N/A & 32.15 & 14.82 & N/A & $\sim 150$ \\
        % ~ & \our $\pi^{+1}$ (bs 100) & 0 & 4.91 & N/A & 6.70 & 23.15 & N/A & 444 \\
        % ~ & \our $\pi^{+2}$ (bs 10) & 0.1 & 9.82 & N/A & 18.5 & 31.64 & N/A & 300 \\
 \bottomrule    
\end{tabular}
    }
\end{table*}

\section{Experiments}
In this section, we show empirical results of evaluating our method on \dsm and \dsh datasets with problem sizes of $N=20,50,100$. We begin by evaluating the two objectives, legality and optimality, of different models, and contrasting hard and soft constraints. Then, we demonstrate the effect of various components of our methodologies.

\subsection{Setups}
\minisection{Evaluation metrics.}
% evaluate metrics
In order to compare method performance from different perspectives, we use four evaluation metrics: % \textbf{illegal rate} (Illegal), \textbf{solution gap} (Gap), \textbf{solving time} (Time), and \textbf{total timeout} (Timeout).
\begin{itemize}[leftmargin=*]
\item \textbf{Illegal rate} is the proportion of illegal solutions produced by the algorithm in datasets, which reflects the legality of the solutions.
\item \textbf{Solution gap}, \textit{i.e.}, optimality of solutions, is the gap between solution distance $L$ and distance of expert solution, $L_\text{expert}$, calculated by $(L/L_\text{expert}-1)$. In particular, we only calculate the solution gap for legal solutions.
\item \textbf{Solving time}, the execution time each algorithm takes to solve $1\,000$ problem instances. 
\item \textbf{Total timeout}, the sum of timeouts on each node.  
\end{itemize}
All the metrics are evaluated on test sets consisting of $1\,000$ instances.

% evaluation
\minisection{Configuration.}
For fair comparisons, we evaluate solving time on the same hardware configuration. Greedy policies and heuristic policies execute on one Intel(R) Xeon(R) Platinum 8255C CPU @ 2.50GHz (with 8 cores). Learning-based policies execute on one NVIDIA GeForce RTX 3090. We trained the learning-based models for $500\,000$ samples and $100$ epochs. Hyperparameters for training are listed in supplementary materials.
% Appendix~\ref{sec:app_exp}.

\minisection{Baselines.}
We compare three types of baseline algorithms described as follows.
\begin{itemize}[leftmargin=*]
  \item \textbf{Heuristic baseline.} We consider the state-of-art TSPTW solver, \textit{LKH3}~\cite{Helsgaun2017AnEO}, as the oracle to calculate the solution gap and generate datasets. Since we screened out a few instances that LKH3 cannot solve, the illegal rate of LKH3 is 0\%.

  \item \textbf{Greedy baselines.} We provide two trivial rule-based greedy policies as poor baselines.
  Their results may be regarded as fair lower bounds of solver performance. Greedy policies follow the route construction process to generate the solution. At each step, \textit{Minimum time-consuming greedy} (Greedy-MT) chooses the node with the minimum arrival time to visit. The calculation of arrival time includes the waiting time for the earliest access time $t^s_i$. \textit{Minimum latest access time greedy} (Greedy-LT) chooses the node with the minimum latest access time $t^e_i$.

  \item \textbf{Learning-based baselines.} As mentioned in Section~\ref{sec:dataset}, most of the recent learning-based TSPTW approaches are evaluated on inconclusive datasets and did not release available code. We selected two RL works solving similar problems for our comparison. Both works provide open-source code. \textit{JAMPR}~\cite{Falkner2020LearningTS} is the state-of-art for Vehicle Routing Problem with Time Windows (VRPTW), a variant of TSPTW with multiple salesmen. We adapt JAMPR for the TSPTW by removing additional constraints and allowing only one salesman. JAMPR resolves the VRPTW by successively constructing multiple routes and masking all timeout nodes for the current route. In TSPTW, the only route must visit every node without exceeding the timeout limit, and there is no simple, feasible masking that can be used to avoid the timeout.
  The other method is \textit{AM}~\cite{Kool2018AttentionLT}, which is a well-known route construction method. We adapt AM for TSPTW with additional time-window $[t^s_i, t^e_i]$ features as input. 
  The TSPTW solutions adhere to a fixed order, thereby rendering algorithms that rely on solution symmetry, such as POMO~\cite{Kwon2020POMOPO}, inapplicable to the TSPTW problem.
  The total reward function $R$ for RL algorithms consists of route length, timeout period, and the number of timeout nodes,
   \begin{equation*}
    R = R_\text{route\_length} + R_\text{total\_timeout} + R_\text{number\_of\_timeout\_nodes}.
   \end{equation*}

\end{itemize}

\begin{figure*}[htb]
  \centering           
  \subfloat[$N=20$]  
  {
      \label{fig:subfig1}\includegraphics[width=0.3\textwidth]{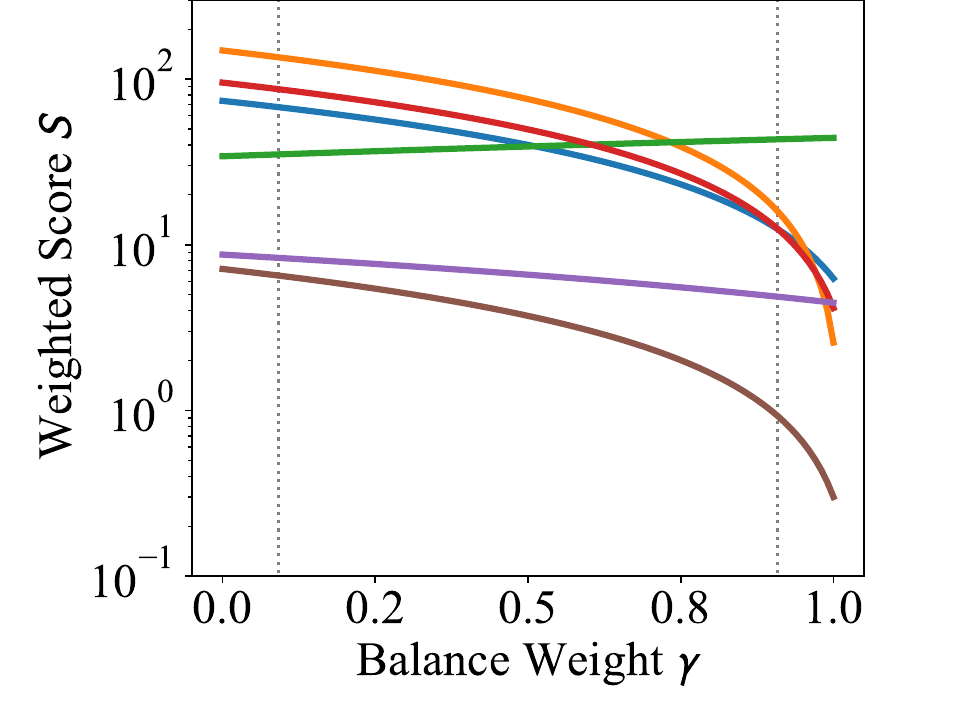}
      
  }
  \subfloat[$N=50$]
  {
      \label{fig:subfig2}\includegraphics[width=0.3\textwidth]{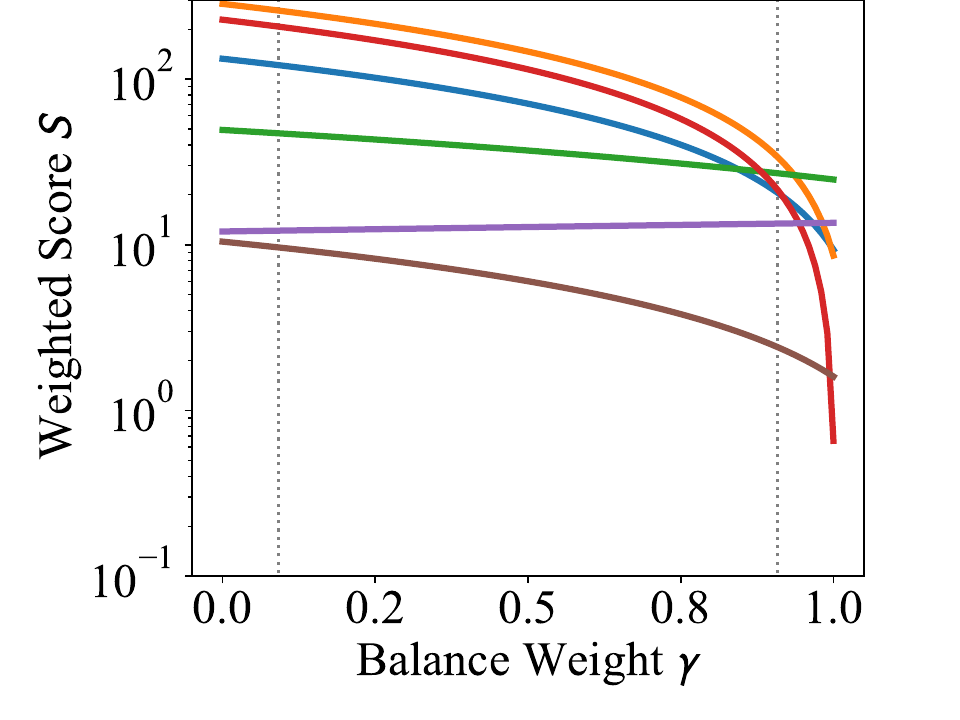}
  }
  \subfloat[$N=100$]
  {
      \label{fig:subfig3}\includegraphics[width=0.3\textwidth]{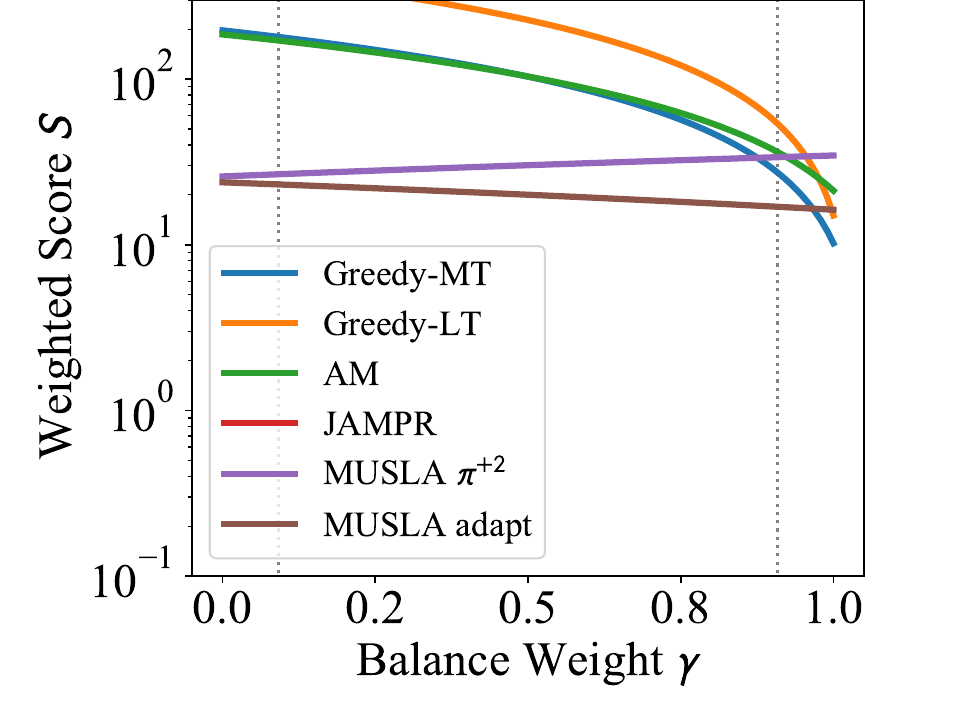}
  }
  \caption{Weighted score of different models. The dotted lines highlight reasonable ranges of $\gamma$. }
  \label{fig:opt}
\end{figure*}

\begin{table*}[htbp]
\centering
\caption{Comparison for mentioned variants of our methods.} \label{table:ablation}
\resizebox{0.8\linewidth}{!}{
\begin{tabular}{c|c|ccc|ccc|c}  
\toprule
 \multicolumn{2}{c|}{\multirow{2}{*}{Methods}} & \multicolumn{3}{c|}{\dsm} & \multicolumn{3}{c|}{\dsh} & \multirow{2}{*}{Time(s)} \\
 \cmidrule{3-8}
 \multicolumn{2}{c|}{}  & Illegal(\%) & Gap(\%)  & \multicolumn{1}{l|}{Timeout} & Illegal(\%) & Gap(\%) & \multicolumn{1}{l|}{Timeout} &  \\ 
 \midrule
\multirow{6}{*}{$N=50$} & Static & 82.70 & 7.60  & 8.56  & 74.30 & 122.07  & 9.87   & 1.16 \\ 
        ~ & Dynamic & 50.30 & 6.16  & 0.64  & 55.60 & 31.41  & 4.12   & 1.36 \\ 
        ~ & OSLA $\pi^{+1}$& 11.80 & 8.15  & 3.53  & 24.50 & 18.55  & 8.43   & 1.56 \\ 
        ~ & \our $\pi^{+2}$  & 8.20  & 7.32  & 1.80  & 18.90  & 16.71  & 4.42  & 7.63   \\ 
        ~ & \our adapt & 0.10  & 5.63  & 0.99  & 3.10  & 15.24  & 2.22  & 45.97  \\ \cmidrule{2-9}
        ~ & OSLA-\dsm & 26.80 & 12.22  & 6.12  & 39.10 & 43.46  & 6.84   & 1.56 \\  \bottomrule
\end{tabular}
}
\vspace{-1em}
\end{table*}

\subsection{Results and Analysis}

Table~\ref{table:results} shows the experimental results on \dsm and \dsh datasets. Overall, on different types and sizes of datasets, \our and \our-adapt outperforms other learning-based baselines by a large margin on the solution gap. 
% Considering the illegal rate, \our and \our-adapt guarantees the best performance in most cases.
However, many of the baselines show extreme imbalances in experiments. While all generated routes are legal, imbalanced policies show a large gap in optimality compared to the expert baseline. In order to better evaluate the ability of the algorithm to balance the two indicators, we use the weighted score to quantify the performance. The weighted score $S$ is calculated by 
\begin{equation}\label{equ:score}
    S = \gamma \cdot \text{Illegal} + (1-\gamma) \cdot \text{Gap},
\end{equation}
where the balance weight $\gamma$ determines the importance of the illegal rate in $S$.
We visualize the weighted score of different algorithms in Figure~\ref{fig:opt} varying the balance rate from $0$ to $1$.

For the two terms $\{\gamma \text{Illegal}, (1-\gamma) \text{Gap}\}$ in the Equation~\ref{equ:score}, $\frac{1}{10}$ to $\frac{10}{1}$ should be a reasonable scale range for ratio $\gamma \text{Illegal}/ (1-\gamma) \text{Gap}$, since the goal of algorithms solving TSPTW is to optimize both of two objectives. We highlight the range with dotted lines. In three different sizes of problems, \our-adpat keeps the lowest score within the reasonable range. Although slightly worse than \textit{Greedy-MT} in problem scale of $N=100$, \our outperforms other baselines in most cases. In previous learning-based works, the weighted score is commonly calculated as $S = \gamma \text{Timeout} + (1-\gamma) \text{Gap}$ with a fixed value of $\gamma$. We show the difference between timeout and illegal rate later.

% 1. greedy and failure  2. show the diff between soft and hard 3. hard dataset
On the \dsm dataset, greedy solutions tend to have high legality and low optimality; as for RL algorithms, a trivial strategy such as \textit{Greedy-LT} is easy to explore but may cause local-optimal issues. 
In our training procedure, we do observe that the RL policy maintains a similar performance as \textit{Greedy-LT} for a long period. 
Although \textit{AM} eventually converges to comparable results in problem sizes of $N=20, 50$, it still shows an imbalance result within a large-scale case. The performance of \textit{JAMPR} is even worse, as it does not improve the optimality much in all cases and totally fails in cases where $N=100$.
% Even if a RL policy,  eventually converges to a comparable result, it still shows more preference for legality. 
 The imbalance between the two objectives shows the backwardness of RL algorithms. 
In contrast, the SL algorithm naturally avoids learning imbalance strategies by imitating high-quality expert datasets. In this way, SL methods learn towards a single objective and can improve both optimality and legality simultaneously.

As mentioned in Section~\ref{sec:dataset}, the training and evaluation instances in the \dsh dataset are sampled from different distributions. Compared with other models, the illegal rate of \textit{AM} increases with a large margin compared with it in $\dsm$, $N=20,50$. The incremented values are $+77.66\%$ and $+39.60\%$ respectively. The increases for \our and \our-adapt are reasonable, which shows the generalizability potential of our method. 
% As mentioned in section~\ref{sec:dataset}, the training and evaluation instances in the \dsh dataset are sampled from different distributions. Therefore, the generalization ability of models has a great influence on their results. Table~\ref{table:generalization} shows the performance drop of different algorithms on problem sizes of $N=50$. The huge drop in illegal metrics explains why RL is more biased towards generating legal solutions on the \dsm dataset but not on the \dsh. Compared with \textit{JAMPR} and \textit{AM}, \our only drop slightly in performance, which shows the generalizability potential of our method. 

% \begin{table}[!ht]
%     \centering
%     \caption{Performence drop between training and evaluation \dsh datasets. The greedy strategy has no generalization issues, so it can be used as a stable baseline in this situation.} \label{table:generalization}
%     \begin{tabular}{|l|lllll|}
%     \hline
%         ~ & Greedy-MT & Greedy-LT & AM & JAMPR & \our \\ \hline
%         Illegal & +10\% & +10\% & +10\% & +10\% & +10\%  \\ 
%         Gap & +10\% & +10\% & +10\% & +10\% & +10\%  \\ \hline
%     \end{tabular}
% \end{table}

On \dsm dataset with a problem size of $N=20$, \our shows a lower illegal rate but a higher total timeout compared with \textit{AM}. Intuitively, minimizing the timeout should also minimize the timeout rate. However, there is a difference between the two objectives in fact.
Replacing the legality indicator illegal rate by total timeout, TSPTW is relaxed as a soft-constrained problem, where solutions can tolerate minor timeouts. The experimental results in Table \ref{table:results} also illustrate the difference between the two types of problems. This counter-intuitive result proves that our method better models the hard-constrained problem, rather than simply tuning the results on the original method.

When comparing LKH3 with \our, it is seen that \our only exhibits shorter solution times for larger problem sizes. Although MUSLA is not capable of completely replacing LKH, it serves as a feasible alternative.
% that offers a favorable balance between time efficiency and solution quality.
For scenarios where rapid response is prioritized, \our presents a viable option, allowing for flexibility in the time-quality trade-off that real-world applications often necessitate.

% \begin{figure}
%   \centering
%   \includegraphics[width=0.7\linewidth]{naive_optimality_100.png} 
%   % \fbox{\rule[-.5cm]{0cm}{4cm} \rule[-.5cm]{4cm}{0cm}}
%   \caption{naive optimally 100} \label{fig:opt}
% \end{figure}

\subsection{Ablation Study}

We conduct ablation experiments on a problem size of $N=50$ to compare the effect of different components in our method. Corresponding to Section~\ref{sec:method}, we set up five different models: \textbf{Static} model is a trivial supervised learning model with only static information of TSPTW as input. \textbf{Dynamic} model adds dynamic information to improve the model's perception of the state of each step in the construction process following Section~\ref{sec:method1}. \textbf{OSLA} introduces the one-step look-ahead mechanism for gathering information on constraint boundaries following Section~\ref{sec:method2}. \textbf{\our} and \textbf{\our-adpat} are the multi-step look-ahead policy described in Section~\ref{sec:method3} and ~\ref{sec:method4}. It is clear that each part of the method improves the performance.

We also show the help of diverse supplement training datasets. \textbf{OSLA-\dsm} is an OLSA model trained using only \dsm dataset. The evident decline in performance indicates that diverse training datasets are necessary.

Compared with the heuristic method, learning-based policies still have a gap in performance. However, with learning-based methods, we can obtain solutions faster at the cost of a slight decrease in performance. According to the results in Table~\ref{table:results}, \our and \our-adapt can achieve a 129x and 19x speedup separately using a single GPU at the problem size of $N=100$. To further reduce the solving time, OSLA, which has a reduced solution time and little performance degradation, may be a viable alternative.

%% file: Conclusion.tex
\section{Conclusion}
  In this paper, we propose a novel and effective solution for a challenging hard-constrained variant of TSP, TSPTW, named multi-step look-ahead (\our). 
  In particular, \our is a supervised-learning method that adopts the looking-ahead information as the feature to improve the legality of TSP with Time Windows (TSPTW) solutions.
  To accurately evaluate and benchmark the statistical performance of various approaches, we also construct TSPTW datasets with hard constraints that can be used by the community to conduct follow-up research.
  With comprehensive experiments, \our demonstrates great performance on diverse datasets, which is far better than existing baselines.
  
  The limitation of our work lies in the requirement for expert datasets, which may be expensive to collect, especially in large-scale cases. It is the major difficulty that prevents us from trying larger-scale problems.
  In the future, we plan to improve the search strategy of \our to collect more critical information while reducing the time-consuming. Utilizing suboptimal datasets generated by RL methods could also be a potential direction to address the data generation issues.

%% file: appendix.tex
\appendix
\section{Appendix}
\subsection{Network Structure} \label{sec:app_net}

% \subsection{Network with Dynamic Encoder}

\begin{figure}[t]
  \centering
  \includegraphics[width=\linewidth]{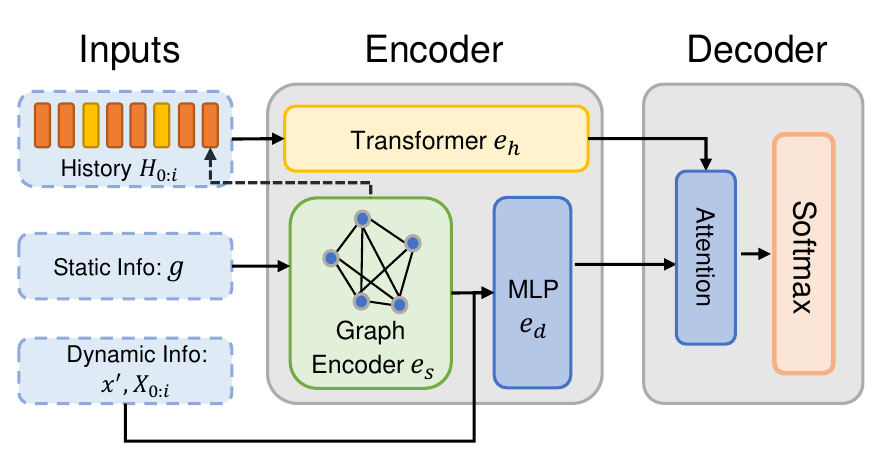}
  % \vspace{-4pt}
  \caption{Network structure of our policy.}\label{fig:network}
  % \vspace{-4pt}
\end{figure}

Following the recent learning-based TSP paper, such as \citet{Kool2018AttentionLT}, the end-to-end policy $\pi_\theta$ consists of an encoder and decoder, parameterized by $\theta$. The encoder takes information needed to build the path as input and produces embedding. Then decoder takes embedding as input and produces the next node $x_i$. Different from recent works\cite{Alharbi2021SolvingTS} which take static information $\{a_i, ts_i, te_i\}$ and history embedding of $X_{0:i}$ as input of policy, we additionally design dynamic node features to describe current solution $X_{0:i}$. Figure~\ref{fig:network} presents the network structure of the policy.

% The dynamic feature of each unvisited node $x' \in V \setminus X_{0:i}$ includes differences between $x_i$ and $x'$ in location and time dimensions. For example, in location dimension, features $\{L_{x', x_i}, a_{x'}-a{x_i}\}$ describe the distance and direction between $x'$ and current node $x_i$. In time dimension, features $\{ts_{x'}-t_i, te_{x'}-t_i\}$ describe the time difference between time windows of $x'$. Details of features design are listed in appendix~\ref{sec:app_feature}.

In order to encode static, dynamic, and historical information, we construct the encoder with three model structures. The first part, the static graph encoder $e_s(\cdot)$, encodes the graph structure of the current TSPTW instance with graph attention networks~\cite{Kaminski2021RossmanntoolboxAD}. The graph encoder aggregates neighbor information across nodes and captures the graph structure of nodes for a specific TSPTW instance $g$. However, the powerful graph neural network structure also brings a huge computational overhead. In order to ensure that encoding dynamic information at each step does not bring too much computational burden, we apply the second part with a node-wise MLP $e_d(\cdot)$ for the dynamic node feature. Given the embedding of encoded static and dynamic features at step $i$ as $h^s_x,h^i_{x}$, the historical sequence of $X_{0:i}$ can be described as $H_{0:i}=\{(h^s_{x_0}, ), (h^s_{x_1}, h^0_{x_1}), \dots, (h^s_{x_i}, h^{i-1}_{x_i})\}$.
As the third part of the encoder, we use a Gated Transformer-XL~\cite{Parisotto2019StabilizingTF} $e_h(\cdot)$ to model the historical embedding sequence. The encoder can be formulated as follows,
\begin{equation}
    h^s = e_s(g),\ h^i_{x'} = e_d(x', h^s,X_{0:i}),\ h^h_i = e_h(H_{0:i}).
\end{equation}

At each construct step $i$, our model predicts the probability of visiting each unvisited node with attention mechanism $d(\cdot)$ and soft-max. The queries come from historical embedding $H_{0:i}$ and the keys and values come from the dynamic embedding $h^i_{x'}$. The formulation for the probability of node $x'$ at step $i$ can be written as
\begin{equation}
  p(x=x'|X_{0:i},g)=\frac{d(h^i_{x'}, h^h_i)}{\sum_{j \in V \setminus X_{0:i}} d(h^i_{j}, h^h_i)},\ x' \in V \setminus X_{0:i}.  
\end{equation}

% Supervised Loss
% Existing the challenge of balancing between legality and optimality, learning from a state-of-art expert rather than exploring strategy by yourself could be a better solution. We construct high-quality datasets with LKH3 solver and sampled TSPTW instances following section~\ref{sec:dataset}.
% The goal of our model $\pi_\theta$ is maximizing the log-probability of expert solutions $X^*$, i.e. imitate the expert policy,
% \begin{equation}
%     \max_\theta \log p_\theta(X=X^* | g)
% \end{equation}

% \subsection{Extend Dynamic Encoder with Multi-step Look-ahead Information}

\subsection{Weakness of Existing Datasets} \label{sec:app_weak_data}
In this section, we demonstrate the weakness of TSPTW instances generated by previous learning-based papers. 
The coordination of TSPTW instances is sampled following uniform distribution, which is consistent with TSP papers. Only the generation of time windows is distinct.
\citet{Alharbi2021SolvingTS,Ma2019CombinatorialOB,Cappart2020CombiningRL} generate time windows following the visiting time $t'_i$ of a constructed solution $X'$. The route $X'$ is constructed following different methods. 

\citet{Alharbi2021SolvingTS,Ma2019CombinatorialOB} construct route $X'$ as a near-optimal TSP solution with learning-based and heuristic solver separately, which provides a too strong prior for the TSPTW instances. The solutions for generated TSPTW instances are basically in line with the solutions of TSP instances, \textit{i.e.} TSPTW solvers can ignore the time window constraints. In order to verify our speculation, we generate a similar dataset, where route $X'$ is constructed by greedily visiting the unvisited nearest neighbor. On this dataset, the \textbf{Greedy-MT} baseline reaches  $\text{Gap}=0.00\%$ and $\text{Illegal}=0.00\%$. \citet{Cappart2020CombiningRL} generate $X'$ with a random permutation, which seems a better choice. However, we test a greedy method that chooses the node $i$ with the minimum earliest access time $t^s_i$, and the results are $\text{Gap}=0.00\%, \text{Illegal}=3.32\%$.

\citet{Zhang2020DeepRL} generate problem instances on a relaxed variant of TSPTW, called TSPTWR. The time windows are generated from a random distribution. However, the constraints are too tight for TSPTW. We generate problem instances following this paper and solve them with LKH3, only $1\%$ instances can find solutions.

% \subsection{More Details}
% Due to page limitations, hyperparameters and comprehensive explanations of the feature and dataset designs, as well as hyperparameters, are included in the \textbf{supplementary materials}.

% \bibliographystyle{plainnat}
% \bibliography{bibfile}
% \printbibliography[heading=subbibliography]

%% file: supplement.tex
% \section{Supplementary}
\subsection{Datasets} \label{sec:app_data}
In this paper, we generate TSPTW datasets with sizes of $n=20, 50, 100$. For small problem sizes $n=20, 50$, the training datasets consist of $500\,000$ problem instances. For big problem size $n=100$, we only generated $50\,000$ TSPTW instances due to the long solution time of LKH3. The data volume ratio for the \dsm, \dsh, and supplementary training sets is $1:1:3$. We generate \dsm with hyperparameter of $\alpha=0.5, \beta=0.75$.
\subsubsection{Hard Dataset}
\paragraph{Training data.} 
The training data of \dsh samples time windows is based on the random distribution of \dsm. For a TSPTW instance $g$ with the size of $n$, we select $\lfloor 0.3n \rfloor$ nodes and divide them into $n_g$ groups. Each group of nodes regenerates time windows based on the random distribution of \dsm and adds an offset time on the time windows. More specifically, the generation process consists of the following four steps.
\begin{enumerate}
    \item Sample time windows following \dsm dataset.
    \item Randomly pick $\lfloor 0.3n \rfloor$ nodes and divide them into $k_p$ groups. For a problem size of $n=20$, the total number of groups is $k_p=2$. For problem sizes of $n=50, 100$, the total number of groups is sampled from a range, $k_p\sim \mathcal{U}[2,7]$.
    \item Individually resample time windows for each group following distribution of \dsm. In particular, the parameter $T_n$ for group with $n_p$ nodes is $T_{n=n_p}$. 
    \item For a group $p$, sample a start time $t_p$ from a uniform distribution, $t_p \sim \mathcal{U}[0, T_n]$, then time windows of all nodes in this group is shifted with value $t_p$. 
    \begin{equation}
        t^s_i \gets t^s_i + t_p
        ,\quad t^e_i \gets t^e_i + t_p
    \end{equation}
    
\end{enumerate}
\paragraph{Evaluation data.}

Evaluation \dsh data are generated using the same method of training \dsh data. The generation process of evaluation data is modified in steps 1 and 3. At step 1, time windows are constant values as $t^s = 0, t^e=T_n$. At step 3, time windows are constant values as $t^s = 0, t^e = T_{n=n_p}$. 

\subsubsection{Supplementary Training Datasets}
\paragraph{Weakly constrained data.} 
\dsm and \dsh datasets are designed to have conflicts in legality and optimality, which makes the learning-based policy are tends to generate legal solutions with poor performance. Therefore, we added two weakly constrained datasets based on \dsm. The first dataset removes the constraint of earliest accessing time, \textit{i.e.} $t^s_i = 0$. The second dataset removes both sides of time windows constraints, \textit{i.e.} $t^s_i = t^e_i = 0$. For the second dataset, the TSPTW instances are relaxed to TSP instances.

\paragraph{Grouped \dsm data.} Grouped \dsm data is designed as a supplementary of \dsm and \dsh. Similar to \dsh training data, the grouped \dsm data divided all $n$ nodes into $k_p$ groups. The shift value $t_p$ of time windows in i-th group $p$ is the maximum value of $t^e$ in the (i-1)-th group, which ensures that time windows from two different groups do not cover each other.
\subsection{Feature Design} \label{sec:app_feature}
\subsubsection{Static Features}
Static features are designed to describe specific TSPTW instance $g$. Problem instance $g$ is a complete graph $(V, E)$ consisting of the node set $V$ and the edge set $E$. The properties of nodes are described with static node features, which are listed in Table~\ref{table:static_node_feature}.
The properties of edges are described with static edge features and are listed in Table~\ref{table:static_edge_feature}. In order to reduce the computational burden, we only retain the top $20\%$ nearest neighbors to add edge features for each node.
% In order to ensure enough information for imitating the expert policy, we have included as many different features as possible, even if some information is redundant. It does not impose too much additional burden on computational. 

\begin{table*}[!ht]
    \centering
    \caption{Static node features for node $i$.} \label{table:static_node_feature}
    \resizebox{0.8\linewidth}{!}{
    \begin{tabular}{l|c|l}
    \toprule
        Description & Feature & Dimension \\ \midrule
        node coordination & $a_i$ & 2  \\ 
        time windows &  $\{t^s_i, t^e_i\}$ & 2  \\ 
        difference in coordinates between node $i$ and starting node $0$ &   $a_i-a_0$ & 2  \\
        distance to starting node $0$ & $L_{i,0}$ & 1  \\ 
        \bottomrule
    \end{tabular}
    }
\end{table*}

\begin{table*}[!ht]
    \centering
    \caption{Static edge features for edge $(i, j)$.} \label{table:static_edge_feature}
    
    \resizebox{0.75\linewidth}{!}{
    \begin{tabular}{l|c|l}
    \toprule
        Description & Feature & Dimension \\ \midrule
        distance between nodes $i,j$ & $L_{i,j}$ & 1  \\ 
        \multirow{2}{*}{difference in time windows between node $i,j$} & $\{t^s_j-t^s_i, t^e_j-t^s_i,$ & \multirow{2}{*}{4}  \\  
        ~ & $\quad t^e_j-t^s_i, t^e_j-t^e_i\}$ & ~ \\  
        \bottomrule
    \end{tabular}
    }
\end{table*}

\subsubsection{Dynamic Features}
Dynamic features are designed to describe all unvisited nodes $x'\in V\setminus X_{0:i}$ given a specific TSPTW instance $g$ and current partial tour $X_{0:i}$. Dynamic features for node $x'$ are listed in Table~\ref{table:dynamic_feature}, where $t_i$ is the current time for partial tour ${X_{0:i}}$.

\begin{table*}[!ht]
    \centering
    \caption{Dynamic features for node $x'$ at step $i$ of route construction.} \label{table:dynamic_feature}
    \resizebox{0.8\linewidth}{!}{
    \begin{tabular}{l|c|l}
    \toprule
        Description & Feature & Dimension \\ \midrule
        node coordination & $a_{x'}$ & 2  \\ 
        difference in coordinates between node $i$ and next node $x'$ &   $a_{x'}-a_{x_i}$ & 2  \\
        distance from current node $x_i$ to next node $x'$ & $L_{x', x_i}$ & 1  \\ 
        time spent visiting node $x'$ &  $\max(L_{x', x_i}+t_i, t^s_{x'})-t_i$ & 1  \\ 
        time difference between time windows of node $x'$ at step $i$ &  $\{t^s_{x'}-t_i, t^e_{x'}-t_i\}$ & 2  \\ 
        \multirow{2}{*}{difference in time windows between node $i,j$} & $\{t^s_{x'}-t^s_{x_i}, t^e_{x'}-t^s_{x_i},$ & \multirow{2}{*}{4}  \\  
        ~ & $\quad t^s_{x'}-t^e_{x_i}, t^e_{x'}-t^e_{x_i}\}$ & ~ \\  
        \bottomrule
    \end{tabular}
    }
\end{table*}

\subsubsection{Look-ahead Node Features}
Following Section 4, we construct a look-ahead route $X'$ for node $x'$ and gather the look-ahead information $I^d_{+1}$ or $I^d_{+2}$ as look-ahead node features of $x'$. The look-ahead information $I^d_{+1}$ and $I^d_{+2}$ have the same feature design where only the way of constructing $X'$ is different. The look-ahead features consist of two parts, constraint violation features and greedy features. For simplicity, we introduce the one-step look-ahead features. With partial route $X'$, $x'$ is the current node, and $t_{i+1}$ is the current time.

The constraint violation features describe the possible delay caused by the current route $X'$ for time constraint $t^e$. We denote the set of nodes that are already to be late as $X''_\text{late}=\{x''\in V\setminus X' | t_{i+1} + L_{x', x''} > t^e_{x''}  \}$. 
The detailed feature design is as follows.
\begin{itemize}
    \item Feature $f^d_1$ denotes if the set $X''_\text{late}$ is an empty set. If $X''_\text{late} \neq \varnothing$, route $X'$ definitely cause a timeout.
    \begin{align}
        f^d_1 = 
        \begin{cases}
            1,\quad &|X''_\text{late}| > 0 \\
            0,\quad &|X''_\text{late}| = 0
        \end{cases} 
    \end{align}
    
    \item Feature $f^d_2, f^d_3$ represents the degree to which the current route $X'$ violates constraints. 
    \begin{align}
        f^d_2 &= \max_{x''\in X''_\text{late}} t_{i+1} + L_{x', x''}  - t^e_{x''} \\
        f^d_3 &= \sum_{x''\in X''_\text{late}} t_{i+1} + L_{x', x''}  - t^e_{x''} 
    \end{align}

\end{itemize}

For greedy features, we select the unvisited node $x''_g$ with the lowest time overhead, $x''_g = \argmin_{x''\in V\setminus X'} \max(L_{x'', x'}+t_{i+1}, t^s_{x''})$. The greedy features $f^d_4, f^d_4$ are the distance and time overhead to node $x''_g$.
\begin{align}
    f^d_4 &= L_{x', x''_g} \\
    f^d_5 &= \max(L_{x', x''_g}+t_{i+1}, t^s_{x''_g}) - t_{i+1}
\end{align}

In addition, the look-ahead information for some nodes is not gathered. For example, the visited nodes and nodes that are not searched in \our do not have meaningful information $I^d$. We add an indicator $f^d_6=\{0,1\}$ to indicate whether the look-ahead information of the node $x'$ exists.

\subsection{Experiment Hyperparameters} \label{sec:app_exp}
Table~\ref{table:hyperparam} lists the common \our parameters used in the experiments.

\begin{table*}[!htbp]
    \centering
    \caption{\our Hyperparameters} \label{table:hyperparam}
    \resizebox{0.8\linewidth}{!}{
    \begin{tabular}{l|l}
    \toprule
        Parameter & Value \\ \hline
        optimizer & AdamW \cite{Loshchilov2017DecoupledWD}  \\ 
        number of hidden units per layer & 128  \\ 
        % optimizer & +10\%  \\ 
        % graph neural network (GNN) & EGAT \\
        number of hidden layers in dynamic encoder & 3 \\
        number of hidden layers in history encoder & 3 \\
        number of hidden layers in static encoder & 5 \\
        nonlinearity & ReLU \\
        normalization & Layer Normalization \cite{Ba2016LayerN} \\
        learning rate & 0.001  \\ \bottomrule
    \end{tabular}
    }
\end{table*}